\documentclass{article}

\usepackage{microtype}
\usepackage{graphicx}
\usepackage{booktabs} 
\usepackage[none]{hyphenat}
\usepackage{amsmath}
\usepackage{amssymb}
\usepackage{floatrow}
\usepackage{float}
\usepackage[caption = false]{subfig}
\usepackage{adjustbox}

\usepackage{hyperref}

\newfloatcommand{capbtabbox}{table}[][\FBwidth]

\usepackage[accepted]{icml2020}

\icmltitlerunning{Variational Prototype Replays for Continual Learning}

\begin{document}

\twocolumn[
\icmltitle{Variational Prototype Replays for Continual Learning}




\begin{icmlauthorlist}
\icmlauthor{Mengmi Zhang}{1,2}
\icmlauthor{Tao Wang}{2}
\icmlauthor{Joo Hwee Lim}{3}
\icmlauthor{Gabriel Kreiman}{1}
\icmlauthor{Jiashi Feng}{2}
\end{icmlauthorlist}

\icmlaffiliation{1}{Boston Children's Hospital, Harvard Medical School}
\icmlaffiliation{2}{National University of Singapore}
\icmlaffiliation{3}{Institute for Infocomm Research, A*STAR, Singapore}

\icmlcorrespondingauthor{Mengmi Zhang}{Mengmi.Zhang@childrens.harvard.edu}

\icmlkeywords{Continual learning, catastrophic forgetting}

\vskip 0.3in
]



\printAffiliationsAndNotice{} 

\begin{abstract}
Continual learning refers to the ability to acquire and transfer knowledge without catastrophically forgetting what was previously learned. In this work, we consider \emph{few-shot} continual learning in classification tasks, and we propose a novel method, Variational Prototype Replays, that efficiently consolidates and recalls previous knowledge to avoid catastrophic forgetting. In each classification task, our method learns a set of variational prototypes with their means and variances, where embedding of the samples from the same class can be represented in a prototypical distribution and class-representative prototypes are separated apart. To alleviate catastrophic forgetting, our method replays one sample per class from previous tasks, and correspondingly matches newly predicted embeddings to their nearest class-representative prototypes stored from previous tasks. Compared with recent continual learning approaches, our method can readily adapt to new tasks with more classes without requiring the addition of new units. Furthermore, our method is more memory efficient since only class-representative prototypes with their means and variances, as well as only one sample per class from previous tasks need to be stored. Without tampering with the performance on initial tasks, our method learns novel concepts given a few training examples of each class in new tasks.
\end{abstract}

\vspace{-8mm}
\section{Introduction}
Continual learning enables humans to continually acquire and transfer new knowledge across their lifespans while retaining previously learnt experiences \cite{hassabis2017neuroscience}. This ability is also critical for artificial intelligence (AI) systems to interact with the real world and process continuous streams of information \cite{thrun1995lifelong}. However, the continual acquisition of incrementally available data from non-stationary data distributions generally leads to catastrophic forgetting in the system \cite{mccloskey1989catastrophic,ratcliff1990connectionist,french1999catastrophic}. Continual learning remains a long-standing challenge for deep neural network models since these models typically learn representations from stationary batches of training data and tend to fail to retain good performance in previous tasks when data become incrementally available over tasks \cite{kemker2018measuring,maltoni2019continuous}.
	
Numerous methods for alleviating catastrophic forgetting have been proposed. The most pragmatical way is to jointly train deep neural network models on both old and new tasks, which demands a large amount of resources to store previous training data and hinders learning of novel data in real time. Another option is to complement the training data for each new task with ``pseudo-data'' of the previous tasks~\cite{shin2017continual,robins1995catastrophic}. In this approach, a generative model is trained to generate fake historical data used for pseudo-rehearsal. Deep Generative Replay (DGR) \cite{shin2017continual} replaces the storage of the previous training data with a Generative Adversarial Network to synthesize training data on all previously learnt tasks. These generative approaches have succeeded over very simple and artificial inputs but they cannot tackle more complicated inputs \cite{atkinson2018pseudo}. Moreover, to synthesize the historical data reasonably well, the size of the generative model is usually very large and expensive in terms of memory resources \cite{wen2018few}. An alternative method is to store the weights of the model trained on previous tasks, and impose constraints of weight updates on new tasks~\cite{he2018overcoming,kirkpatrick2017overcoming,zenke2017continual,lee2017overcoming,lopez2017gradient}. For example, Learning Without Forgetting (LwF) \cite{li2018learning} has to store all the model parameters on previously learnt tasks, estimates their importance on previous tasks and penalizes future changes to these parameters on new tasks. However, selecting the ``important'' parameters for previous tasks via pre-defined thresholds complicates the implementation by exhaustive hyper-parameter tuning. In addition, state-of-the-art neural network models often involve millions of parameters and storing all network parameters from previous tasks does not necessarily reduce the memory cost \cite{wen2018few}. In contrast with these methods, storing a small subset of examples from previous tasks and replaying the ``exact subset'' substantially boost performance~\cite{kemker2017fearnet,rebuffi2017icarl,nguyen2017variational}. To achieve the desired network behavior on previous tasks, incremental Classifier and Representation Learner (iCARL)~\cite{rebuffi2017icarl} follows the idea of logits matching or knowledge distillation in model compression~\cite{ba2014deep,bucilua2006model,hinton2015distilling}. Such approaches rely too much on small amount of data, which easily results in overfitting. In contrast, our method improves generalization by learning class prototypical distributions with their means and variances in the latent space, which can generate multiple samples during replays.
	
In this paper, we propose a method that we call Variational Prototype Replays, for continual learning in classification tasks. Extending previous work~\cite{snell2017prototypical}, we use a neural network to learn class-representative variational prototypes with their means and variances in a latent space and classify embedded test data by finding their nearest representations sampled from class-representative variational prototypes. To prevent catastrophic forgetting, our method replays one sample per class from previous tasks, and correspondingly matches newly predicted representations to their nearest class prototypes stored from previous tasks. Since not all prototypical features learnt from the previous tasks are equally important in new tasks, the learnt variance in variational prototypes of previous tasks provide confidence levels of learnt prototype features, and therefore our method can selectively forget under-represented features in the prototypes while the network learns to adapt to new tasks. We evaluate our method under two typical experimental protocols, incremental domain and incremental class, for few-shot continual learning across three benchmark datasets, MNIST~\cite{deng2012mnist}, CIFAR10~\cite{krizhevsky2009learning} and miniImageNet~\cite{deng2009imagenet}. Compared with  state-of-the-art performance, our method significantly boosts the performance of continual learning in terms of memory retention capability while being able to generalize to learn new concepts and adapt to new tasks, even with a few training examples in new tasks. Unlike parameter regularization methods, our approach further reduces the memory storage by storing only one sample per class as well as variational prototypes in the previous tasks. Moreover, in contrast to methods where the last layer in traditional classification networks often structurally depends on the number of output classes, our method maintains the same network architecture and does not require adding new units.

\vspace{-4mm}
\section{Few-shot Continual Learning Protocols}\label{sec:taskprotocols}

\begin{figure*}[t]
		\begin{center}
			\includegraphics[width=17cm]{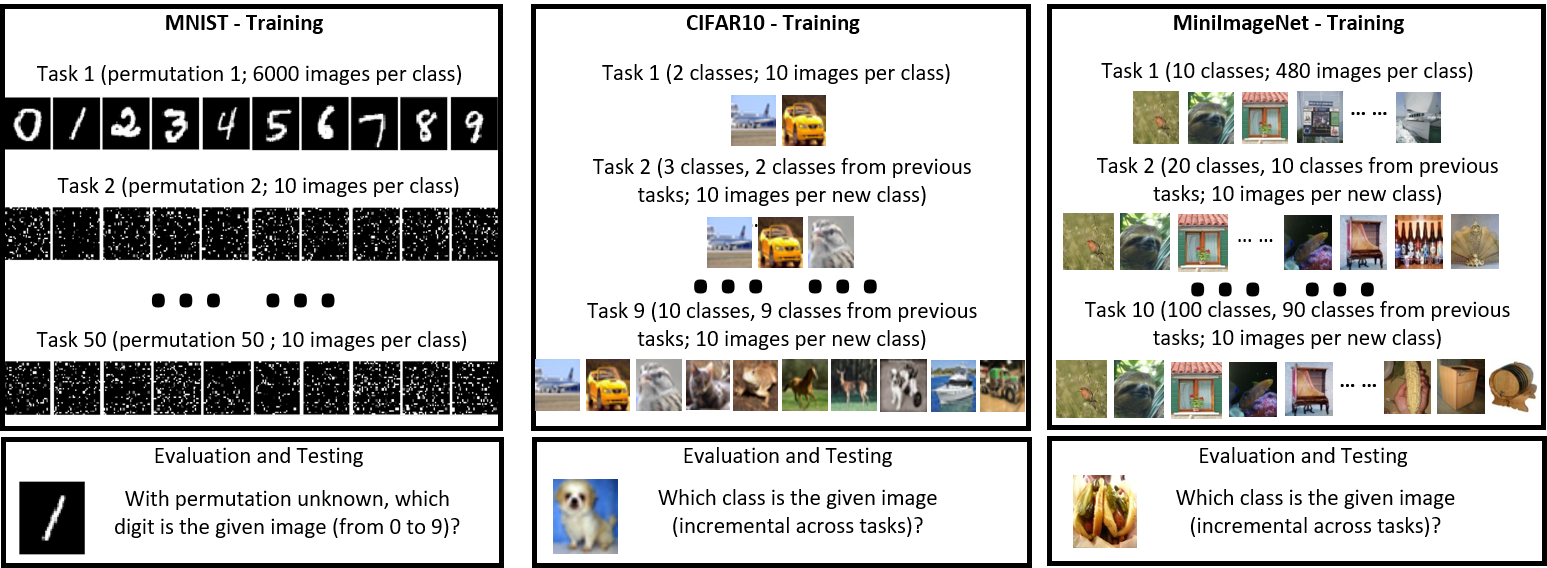}
		\end{center}
		\caption{Schematics of three task protocols in our few-shot continual learning experiments: learning with incremental domain on permuted MNIST; and (b) learning with incremental classes on split CIFAR10 and MiniImageNet. The number of images per class per task is illustrated in the schematics.\vspace{-3mm}}\vspace{-3mm}
		\label{fig:taskdomainintro}
\end{figure*}

Humans can learn novel concepts \textit{given a few examples} without sacrificing classification accuracy on initial tasks \cite{gidaris2018dynamic}. However, typical continual learning schemes assume that a large amount of training data over all tasks is always available for fine-tuning networks to adapt to new data distributions, which does not always hold in practice. We revise task protocols to more challenging ones: networks are trained with a few examples per class in sequential tasks except for the first task in the sequence. For example, we train the models with 6,000 and 480 example images per class in the first task respectively on MNIST and miniImageNet and 10 images per class in subsequent tasks. We also evaluate an even more challenging protocol when there are only 10 example images per class even in the first task in CIFAR10.

\textbf{Permuted MNIST in incremental domain task} is a benchmark task protocol in continual learning~\cite{lee2017overcoming,lopez2017gradient,zenke2017continual} (Figure~\ref{fig:taskdomainintro}). In each task, a fixed permutation sequence is randomly generated and is applied to input images in MNIST~\cite{deng2012mnist}. Though the input distribution always changes across tasks, models are trained to classify 10 digits in each task and the model structure is always the same. There are 50 tasks in total. During testing, the task identity is not available to models. The models have to classify input images into 1 out of 10 digits.
	
\textbf{Split CIFAR10 and split MiniImageNet in incremental class task} is a more challenging task protocol where models need to infer the task identity and at the same time solve each image classification task. The input data is also more complex, including classification on natural images in CIFAR10~\cite{krizhevsky2009learning} and miniImageNet~\cite{deng2009imagenet}. The former contains 10 classes and the latter consists of 100 classes. In CIFAR10, the model is first trained with 2 classes and later by adding one more class in each subsequent task. There are 9 tasks in total and 10 images per class in the training set. In miniImageNet, models are trained with 10 classes in each task. There are 10 tasks in total.
	
\begin{figure*}[t]
		\begin{center}
			\includegraphics[width=17cm, height=9cm]{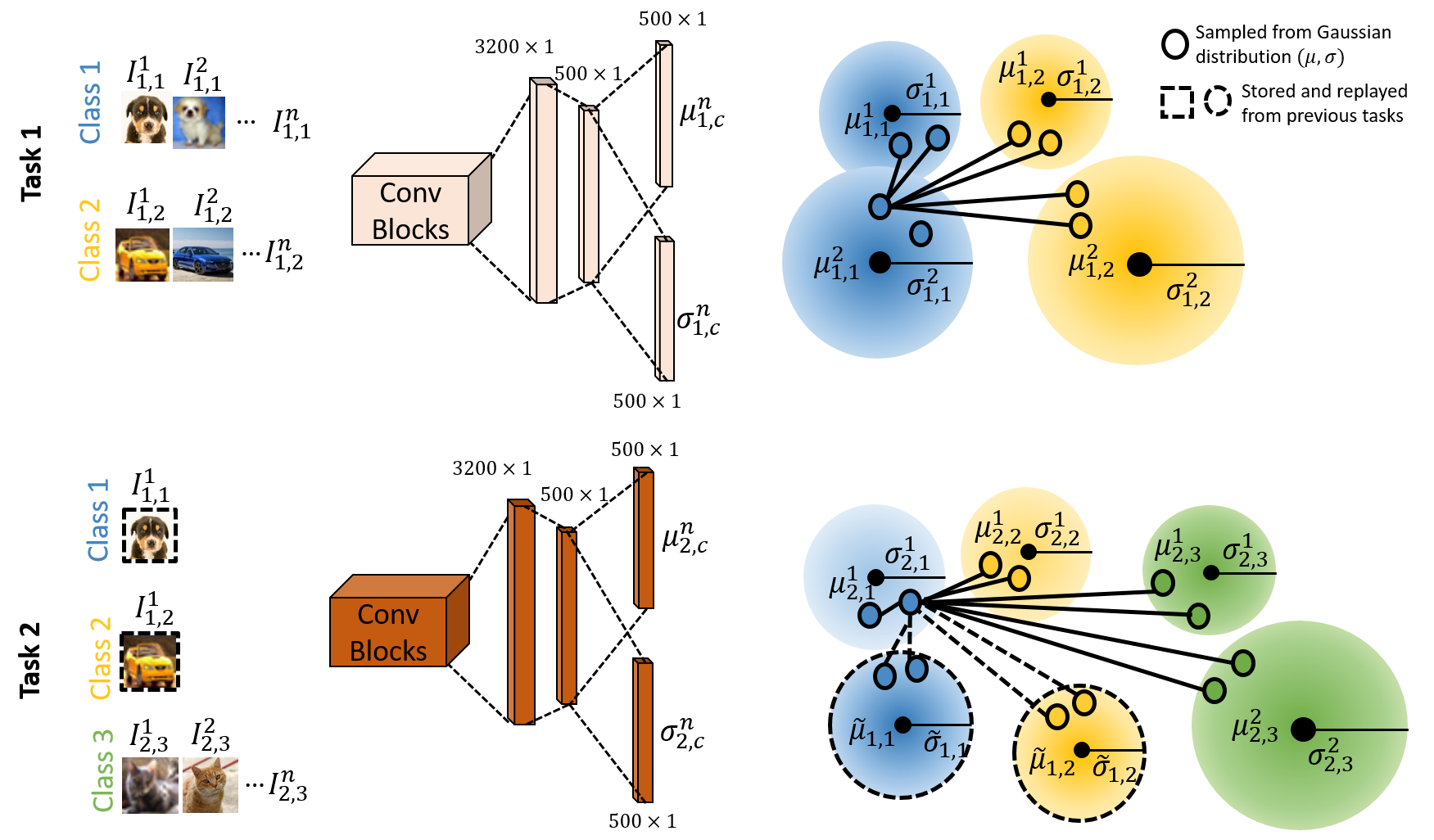}
		\end{center}
		\caption{Illustration of classification in Task 1 and catastrophic forgetting alleviation in Task 2 using our proposed method in the Split CIFAR10 incremental class protocol. In Task 1, there are two classes (blue and yellow) with each class containing $n$ training samples (see Sec~\ref{sec:model} for variable naming conventions). Each training image $I_{1,c}^n$ inputs to a feed-forward 2D-CNN and outputs two vectors: mean $\mu_{1,c}^n$ and variance $\sigma_{1,c}^n$. Their output dimension is $1\times 500$. Multiple samples (solid lined circles) can be generated from gaussian distribution based on each pair of mean $\mu_{1,c}^n$ and variance $\sigma_{1,c}^n$. The color of spheres denotes object class. Classification is performed by comparing the L2-norm distance between any pairs of samples from the same class or different classes. The inter-class distance pairs should be smaller than intra-class ones. Class-representative variational prototypes denoted by prototypical mean $\tilde{\mu}_{1,c}^n$ and prototypical variance $\tilde{\sigma}_{1,c}^n$ for Task 1 are computed by averaging all training samples of the same class. In Task 2, a new class is introduced. The same 2D-CNN architecture is inherited from Task 1 but the parameters of the 2D-CNN get optimized. Only one sample image per class (dash lined square) from Task 1 is replayed. For $I_{1,1}^1$, the new mean $\mu_{2,1}^1$ and variance $\sigma_{2,1}^1$ is computed in Task 2 and similarly we get a new pair of $\mu_{2,2}^1$ and $\sigma_{2,2}^1$ for $I_{1,2}^1$. The classification among three classes can be performed as described in Task 1 (solid straight lines). To eliminate catastrophic forgetting, our method constantly regresses the mean and variance of replayed samples to be as close as possible (dashed straight line) to the class-representative variational prototypes, $\tilde{\mu}_{1,c}^n$ and $\tilde{\sigma}_{1,c}^n$ denoted in dash lined circles, in Task 1.\vspace{-4mm}}\vspace{-8mm}
		\label{fig:model}
\end{figure*}

\vspace{-4mm}
\section{Method}\label{sec:model}

We propose a novel method, Variational Prototype Replays, for few-shot continual learning. First, we introduce variable naming conventions and the problem formulation. Up to any task $t$ where $t \in \{1, 2, ..., T\}$ and $T$ is not pre-determined, there is a  total of $C$ classes, and we use $c$ to denote any class $c \in \{1, 2, ..., C\}$. In the incremental domain protocol, $C=10$ for task $t$; whereas in the incremental class protocol, $C$ increases with the number of tasks. There is a total of $N$ training samples per class and we use $n$ to denote any training sample in a class. To explicitly define a training sample $I_{t,c}^n$, we use superscript to denote the $n$th training sample. For example, $I_{1,2}^3$ denotes the $3$rd training sample from the $2$nd class in the $1$st task. Next, we illustrate how to apply our method to perform classification in a task and how to prevent catastrophic forgetting across tasks (Fig~\ref{fig:model}).
\vspace{-2mm}
\subsection{Classification}\label{subsec:classification}

Our method can be applied on any feed-forward 2D-ConvNet (2D-CNN) architecture for classification tasks. The network with parameters $F_t$ learns to encode an input image $I_{t,c}^n$ in a latent space, in which these encoded image representations cluster around a prototype for each class and classification is performed by finding the nearest prototype (Fig.~\ref{fig:model}). Extending previous work~\cite{snell2017prototypical} on learning a single prototype for each object class $c$ in task $t$, we introduce variational prototypes that follow a Gaussian distribution parameterized with mean $\tilde{\mu}_{t,c}$ and variance $\tilde{\sigma}_{t,c}$. The mean and variance allow the network to replay many prototypes sampled from class-representative distributions to prevent overfitting and allow easy interpolation in the latent space. Compared with other replay methods, such as~\cite{rebuffi2017icarl},  where latent representations of each individual image have to be stored for replays, variational prototypes provide advantages in memory usage since only the mean and variance need to be stored for each class.

 Inspired by the design of variational autoencoders \cite{doersch2016tutorial}, we propose variational encoders which learn a conditional class-representative Gaussian distribution with its mean $\mu^n_{t,c}$ and variance $\sigma^n_{t,c}$, given each input image $I_{t,c}^n$:
 $(\mu^n_{t,c}, \sigma^n_{t,c}) = F_t(I^n_{t,c})$. Variational prototypes $(\tilde{\mu}_{t,c}, \tilde{\sigma}_{t,c})$ can then be computed by taking the average of variational image representations conditioned from all input image $I_{t,c}^n$ belonging to class $c$ in task $t$:

\vspace{-4mm}
 \begin{equation}\label{equ:prototype2}
\tilde{\mu}_{t,c} = \frac{1}{N}\sum_{n}^N \mu^n_{t,c}, \tilde{\sigma}_{t,c} =\frac{1}{N} \sum_{n}^N \sigma^n_{t,c}
\end{equation}
	In task $t$, to perform classification on total $C$ classes, the goal is to make each encoded image's representational distribution to be close to the variational prototype distribution within the same class and to be far apart from other variational prototype distributions of different classes. We sample $Z$ latent representations $s^z_{\mu^n_{t,c}, \sigma^n_{t,c}}$ from both image representational distributions and $s^z_{\tilde{\mu}_{t,c}, \tilde{\sigma}_{t,c}}$ from variational prototype distributions. For each $s^z_{\mu^n_{t,c}, \sigma^n_{t,c}}$ from class $c$, the network estimates a distance distribution based on a softmax over distances to all the sampled prototypes of $C$ classes in the latent space:
	\vspace{-2mm}
	\begin{equation}\label{equ:prototype3}
	\begin{aligned}
	p_{F_t} & (c|s^z_{\mu^n_{t,c},\sigma^n_{t,c}})   = \\ & \frac{\exp(-d(s^z_{\mu^n_{t,c}, \sigma^n_{t,c}},s^z_{\tilde{\mu}_{t,c},\tilde{\sigma}_{t,c}}))}{\sum_{c'}^{C} \exp(-d(s^z_{\mu^n_{t,c}, \sigma^n_{t,c}},s^z_{\tilde{\mu}_{t,c'}, \tilde{\sigma}_{t,c'}}))}.
	\end{aligned}
	\end{equation}

	where we define distance function $d(s^z_{\mu^n_{t,c}, \sigma^n_{t,c}},s^z_{\tilde{\mu}_{t,c}, \tilde{\sigma}_{t,c}})$ as the L2-norm between $s^z_{\mu^n_{t,c}, \sigma^n_{t,c}}$ and $s^z_{\tilde{\mu}_{t,c}, \tilde{\sigma}_{t,c}}$.
	
	The classification objective is to minimize the cross-entropy loss $L_{classi}$ with the ground truth class label $c$ via Stochastic Gradient Descent \cite{bottou2010large}: $L_{classi} = -\log p_{F_t} (c|s^z_{\mu^n_{t,c}, \sigma^n_{t,c}})$
	
	Compared to traditional classification networks with a specific classification layer attached in the end, (also see Table~\ref{tab:memallocationCIFAR} for network architecture comparisons between baseline methods and ours), our method keeps the network architecture unchanged while using the nearest prototypical samples in the latent space for classification. For example, in the split CIFAR10 incremental class protocol where the models are asked to classify new classes (see also Sec~\ref{sec:taskprotocols}), traditional classification networks have to expand their architectures by accommodating more output units in the last classification layer based on the number of incremental classes and consequently, additional network parameters have to be added into the memory.
	
	In practice, when $N$ is large, computing $\tilde{\mu}_{t,c}$ and $\tilde{\sigma}_{t,c}$ is costly and memory inefficient during training. Thus, at each training iteration, we randomly sample two complement image subsets for each class: one subset for computing prototypes and the other for estimating the distance distribution. Sampling size $Z$ also influences memory and computation efficiency. In the split CIFAR10 incremental class protocol, we choose $Z=50$ (see Sec.~\ref{subsec:ablation} for analysis on sampling sizes). Our primary choice of the distance function $d(\cdot)$ is L2-norm which has been verified to be effective in \cite{snell2017prototypical}. As introduced in the network distillation literature \cite{hinton2015distilling}, we include a temperature hyperparameter $\tau$ in $d(\cdot)$  and set its value empirically based on the validation sets. A higher value for $\tau$ produces a softer probability distribution over classes.
	
	\subsection{Variational Prototype Replays}
	
For a sequence of tasks $t\in\{1,2,...,T\}$, the goal of the network with parameters $F_T$ is to retain  good classification performance on all $C$ classes after being sequentially trained over $T$ tasks while it is only allowed to carry over a limited amount of information about previous classes $c_{old}$ from the previous $T-1$ tasks. This constraint eliminates the naive solution of combining all previous datasets to form one big training set for fine-tuning the network $F_T$ at task $T$.

To prevent catastrophic forgetting, here we ask the network with parameters $F_T$ to perform classification on both new classes $c_{new}$ and old classes $c_{old}$ by replaying some example images stored from $c_{old}$ together with all training images from $c_{new}$. Intuitively, if the number of stored image samples is very large, the network could re-produce the original encoded image representations for $c_{old}$ by replays, which is our desired goal. However, this does not hold in practice given limited memory capacity. With the simple inductive bias that the encoded image representations of $c_{old}$ can be underlined by class-representative variational prototypes, instead of classifying $c_{old}$ using the newly predicted variational prototypes with mean $\tilde{\mu}_{T,c_{old}}$ and variance $\tilde{\sigma}_{T,c_{old}}$, the network learns to classify $c_{old}$  based on stored old variational prototypes $(\tilde{\mu}_{t,c_{old}}, \tilde{\sigma}_{t,c_{old}})$ over all the previous tasks $t$.

As described in the previous subsection, in order to classify  $I_{T,c_{new}}^n$ among $c_{new}$ and $c_{old}$, the network learns to encode its image representation in the latent space and compare its samples $s^z_{\mu^n_{T,c_{new}}, \sigma^n_{T,c_{new}}}$ with new variational prototypes ($\tilde{\mu}_{T,c_{new}}, \tilde{\sigma}_{T,c_{new}}$) and stored old prototypes ($\tilde{\mu}_{{T-1},c_{old}}, \tilde{\sigma}_{{T-1},c_{old}}$) for $c_{old}$ from previous task $T-1$. In order to classify $I_{T,c_{old}}^n$, it reviews all the previous tasks $t$. In each previous task $t$, our method compares its samples $s^z_{\mu^n_{T,c_{old}}, \sigma^n_{T,c_{old}}}$ with all stored variational prototypes ($\tilde{\mu}_{{t},c_{old}}, \tilde{\sigma}_{{t},c_{old}}$).

There have been some attempts to select representative image examples to store based on different scoring functions \cite{chen2012super,koh2017understanding,brahma2018subset}. However, recent work has shown that random sampling uniformly across classes yields outstanding performance in continual learning tasks \cite{wen2018few}. Hence, we adopt the same random sampling strategy.

From the first task to current task $T$, the network parameters $F_t$ keep updating in order to incorporate new class representations in the latent space. Hence, the variational prototypes of $c_{old}$ constantly change their representations even for the same class. Not all prototypical features learnt from $c_{old}$ in the previous tasks are equally useful in classifying both $c_{new}$ and $c_{old}$.  As a hypothetical example, imagine that in the first task, we use shape and color to classify red squares versus yellow circles. In the second task, when we see a new class of green circles, we realize shape might not be as good a feature as color; hence, we may need to put ``less weight" on the shape features when we compare with nearest prototypes from $c_{old}$. The variance in the variational prototype provides a confidence score of how representative the prototypical features are. A higher variance indicates that the prototype feature distribution is more spread out; and hence, less representative of $c_{old}$ in the latent space. Thus, we introduce $\tilde{\sigma}_{t,c}$-weighted L2-norm when we compute the distances between $s^z_{\mu^n_{T,c_{old}}, \sigma^n_{T,c_{old}}}$ and $s^z_{\tilde{\mu}_{{t},c_{old}}, \tilde{\sigma}_{{t},c_{old}}}$ for all previous tasks $t\in \{1,..,T-1\}$:

\begin{equation*}\label{equ:prototype5}
\begin{aligned}
&p_{F_T}  (c_{old}|s^z_{(\mu^n_{T,c_{old}},\sigma^n_{T,c_{old}})})   = \\ & \frac{\exp(-d(s^z_{(\mu^n_{T,c_{old}}, \sigma^n_{t,c_{old}})},s^z_{(\tilde{\mu}_{t,c_{old}}, \tilde{\sigma}_{t,c_{old}})}, \tilde{\sigma}_{t,c_{old}}))}{\sum_{c'_{old}}^{C_{old}} \exp(-d(s^z_{(\mu^n_{T,c_{old}}, \sigma^n_{T,c_{old}})},s^z_{(\tilde{\mu}_{t,c'_{old}}, \tilde{\sigma}_{t,c'_{old}})},\tilde{\sigma}_{t,c'_{old}}))}.
\end{aligned}
\end{equation*}
where we define the weighted distance function:
\begin{equation}\label{equ:prototype6}
d(s_1^z, s_2^z, \sigma) = \|\exp(-0.5 \sigma) \cdot (s_1^z- s_2^z)\|_2
\end{equation}

\setlength{\textfloatsep}{18pt}
\begin{algorithm}[tb]
\caption{Variational Prototype Replays at current task $T$}
   \label{algo:algorithm1}
\begin{algorithmic}
   \STATE {\bfseries Input:} stored images $I_{T,c_{old}}^n$, stored variational prototypes ($\tilde{\mu}_{{t},c_{old}}, \tilde{\sigma}_{{t},c_{old}}$), new training images $I_{T,c_{new}}^n$, network parameters $F_T$

   \STATE {\bfseries Training:}
   \FOR{batch {\bfseries in} $I_{T,c_{new}}^n$}
   \IF{$T = 1$}
   \STATE Train $F_T(I_{T,c_{new}}^n)$ based on ($\tilde{\mu}_{{T},c_{new}}, \tilde{\sigma}_{{T},c_{new}}$)
   \ELSE
   \STATE Train $F_T(I_{T,c_{new}}^n)$ based on ($\tilde{\mu}_{{T-1},c_{old}}, \tilde{\sigma}_{{T-1},c_{old}}$) and ($\tilde{\mu}_{{T},c_{new}}, \tilde{\sigma}_{{T},c_{new}}$)
   \FOR{$t=1$ {\bfseries to} $T-1$}
   \STATE Train $F_T(I_{T,c_{old}}^n)$ based on $(\tilde{\mu}_{{t},c_{old}}, \tilde{\sigma}_{{t},c_{old}})$
   \ENDFOR

   \ENDIF

   \ENDFOR
   \IF{$T>1$}
   \STATE Compute and store ($\tilde{\mu}_{{T},c_{old}}, \tilde{\sigma}_{{T},c_{old}}$) using $I_{T,c_{old}}^n$
   \ENDIF
   \STATE Compute and store ($\tilde{\mu}_{{T},c_{new}}, \tilde{\sigma}_{{T},c_{new}}$) using $I_{T,c_{new}}^n$

\end{algorithmic}

\end{algorithm}

For replays in new tasks, given a limited memory capacity, our proposed method has to store a small image subset and one variational prototype including its mean and variance for each old class $c$ in all previous tasks $t<T$. When the total number of tasks $T$ is small, the memory can store more image examples per class. Dynamic memory allocation enables more example replays in earlier tasks, putting more emphasis on reviewing earlier tasks which are easier to forget. Pseudocode to our proposed algorithm in split CIFAR10 in the incremental class protocol for a training episode is provided in Algorithm~\ref{algo:algorithm1}. The source code of our proposed algorithm is downloadable: \url{https://github.com/kreimanlab/VariationalPrototypeReplaysCL}.

\section{Experimental Details}

We introduce baseline continual learning algorithms with different memory usage over three task protocols.
\begin{table*}[t]
\footnotesize
\begin{tabular}{|c|c|c|c|c|c|c|}
\hline
\begin{tabular}[c]{@{}c@{}}Network architecture\\(baseline method)\end{tabular}                                                                                      & \multicolumn{6}{c|}{conv(3,20,5)$\rightarrow$ conv(20,50,5)$\rightarrow$ fc(3200,500)$\rightarrow$ fc(500,10)$\rightarrow$ softmax}                            \\ \hline
\begin{tabular}[c]{@{}c@{}}Number of network parameters\\(baseline methods)\end{tabular} & \multicolumn{6}{c|}{$3\times20\times5\times5+20\times50\times5\times5+3200\times500+500\times10 = 16.3 \times 10^5$} \\ \hline
\begin{tabular}[c]{@{}c@{}}Network architecture\\(our method)\end{tabular}                                                                                           & \multicolumn{6}{c|}{conv(3,20,5)$\rightarrow$ conv(20,50,5) $\rightarrow$ fc(3200,500)$\rightarrow$ fc(500,1000)$\rightarrow$ nearest prototype }                                  \\ \hline
\begin{tabular}[c]{@{}c@{}}Number of network parameters\\(our method)\end{tabular}        & \multicolumn{6}{c|}{$3\times20\times5\times5+20\times50\times5\times5+3200\times500+500\times1000 = 21.3 \times 10^5$} \\ \hline

 & \multicolumn{2}{c|}{EWC-online} &  MAS & L2 & SI  & ours  \\  \hline
Memory size ($\times10^5$)                                                                                & \multicolumn{2}{|c|}{\begin{tabular}[c]{@{}c@{}}$16.3\times2$\\=32.6\end{tabular} }              & \begin{tabular}[c]{@{}c@{}}$16.3\times2$\\=32.6\end{tabular}             & \begin{tabular}[c]{@{}c@{}}$16.3\times2$\\=32.6\end{tabular}            & \begin{tabular}[c]{@{}c@{}}$16.3\times2$\\=32.6\end{tabular}             & \begin{tabular}[c]{@{}c@{}}$21.3+0.4$\\=\textbf{21.7}\end{tabular}            \\ \hline

\end{tabular}
\caption{Network architecture and memory allocation for continual learning methods on split CIFAR10 in incremental class task. For simplicity, only the network layers with learnable parameters are presented. Other network specifications, such as paddings, activation layers and pooling layers are omitted here. $conv(x,y,z)$ denotes 2D convolutional layer where $x$ is the input channel number, $y$ is the output channel number, and $z$ is the convolutional kernel size. $fc(x,y)$ denotes full-connected layer where $x$ is the input vector size and $y$ is the output vector size. The number in bold denotes the most efficient memory size. \vspace{-4mm}}\vspace{-4mm}\label{tab:memallocationCIFAR}
\end{table*}

\begin{figure*}[t]
		\centering		
		\subfloat[ Average classification accuracies over total 9 tasks]{\includegraphics[width= 5.5cm, height = 4.7cm]{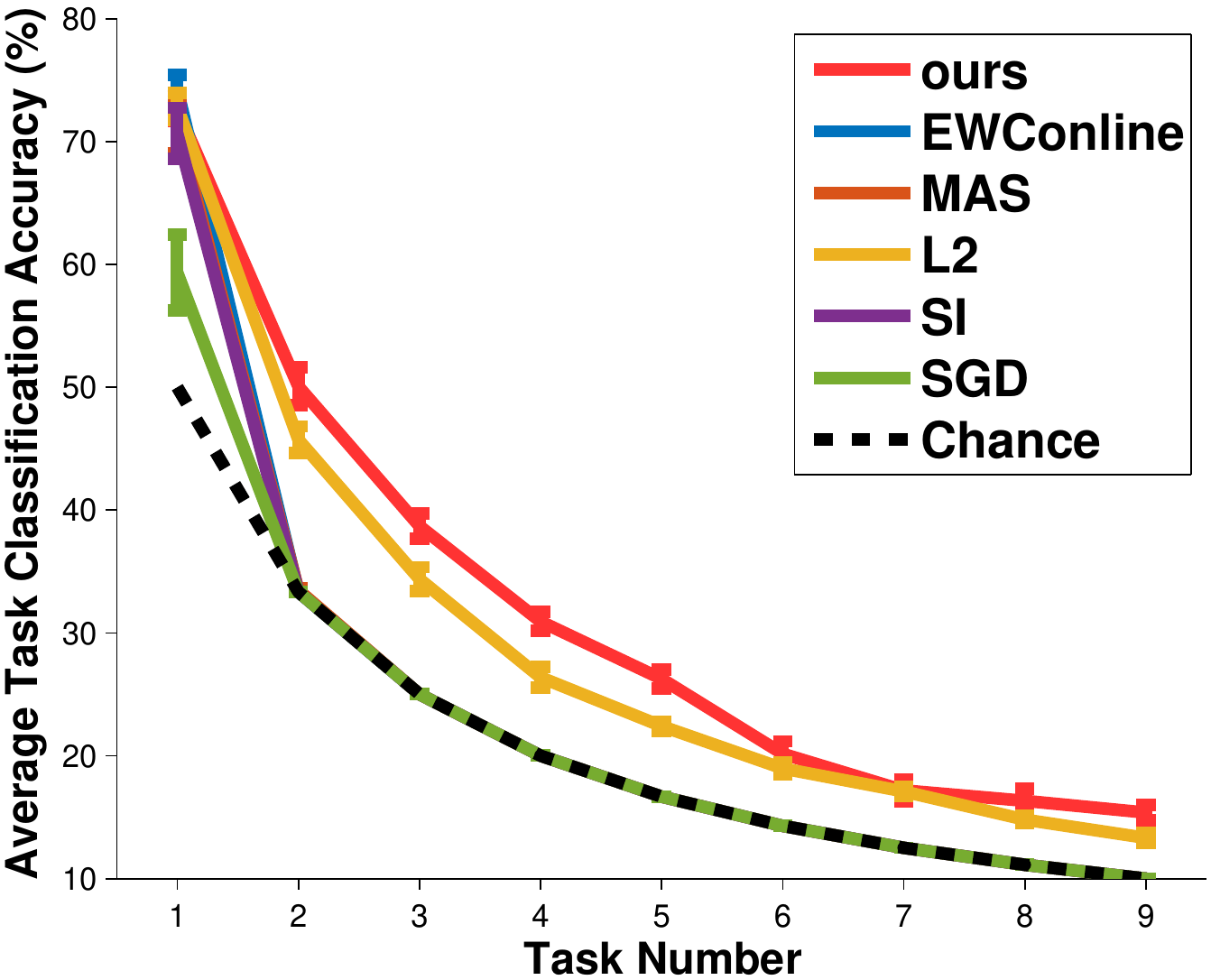}\label{fig:fewshotsavgcifar}}\hspace{0.4cm}
\subfloat[Embedding clusters and prototypes learnt by our method in Task 1]{\includegraphics[width= 5.3cm, height = 4.7cm]{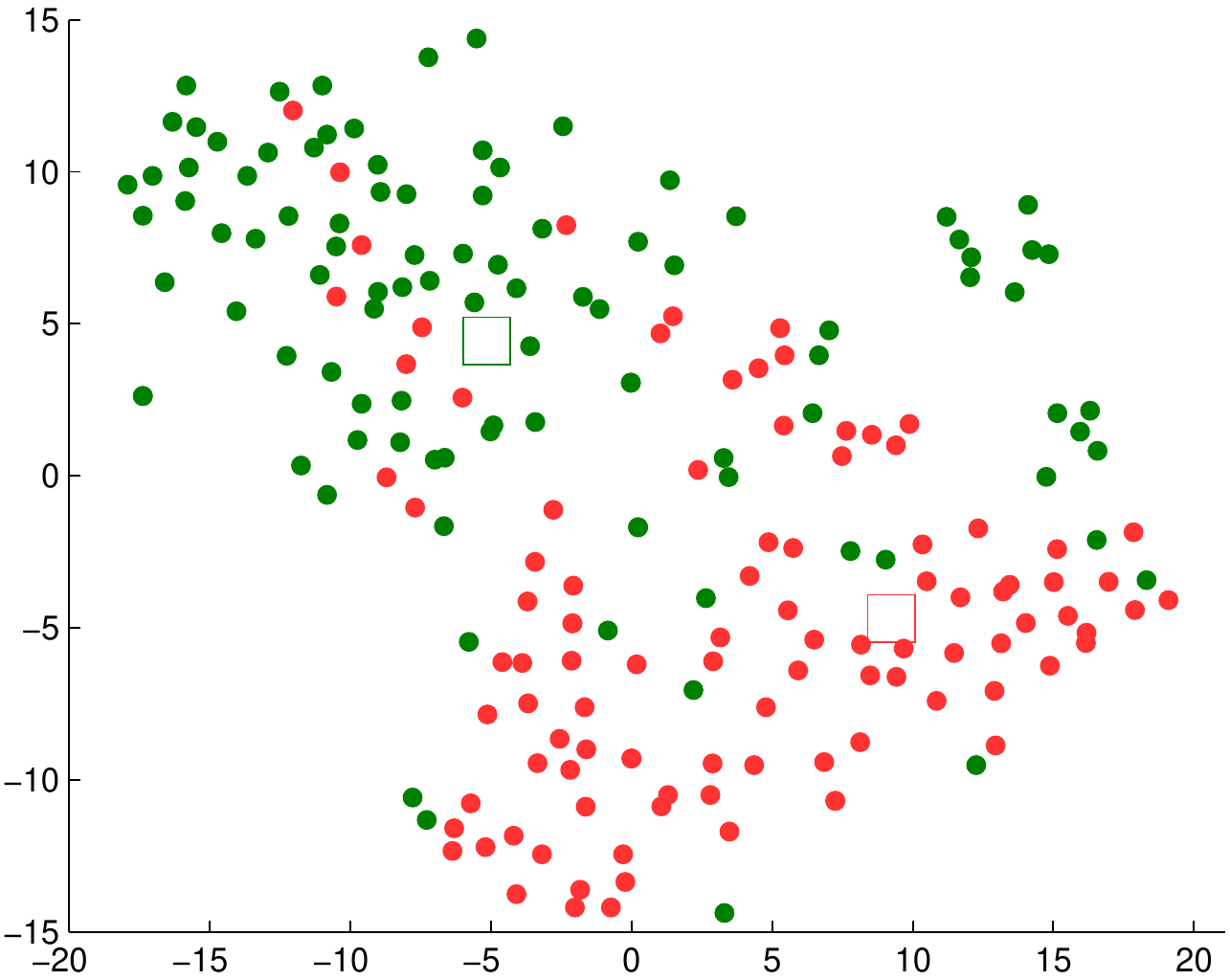}\label{fig:vizclustertask1}}\hspace{0.3cm}
		\subfloat[Embedding clusters and prototypes learnt by our method in Task 3]{\includegraphics[width= 5.3cm, height = 4.7cm]{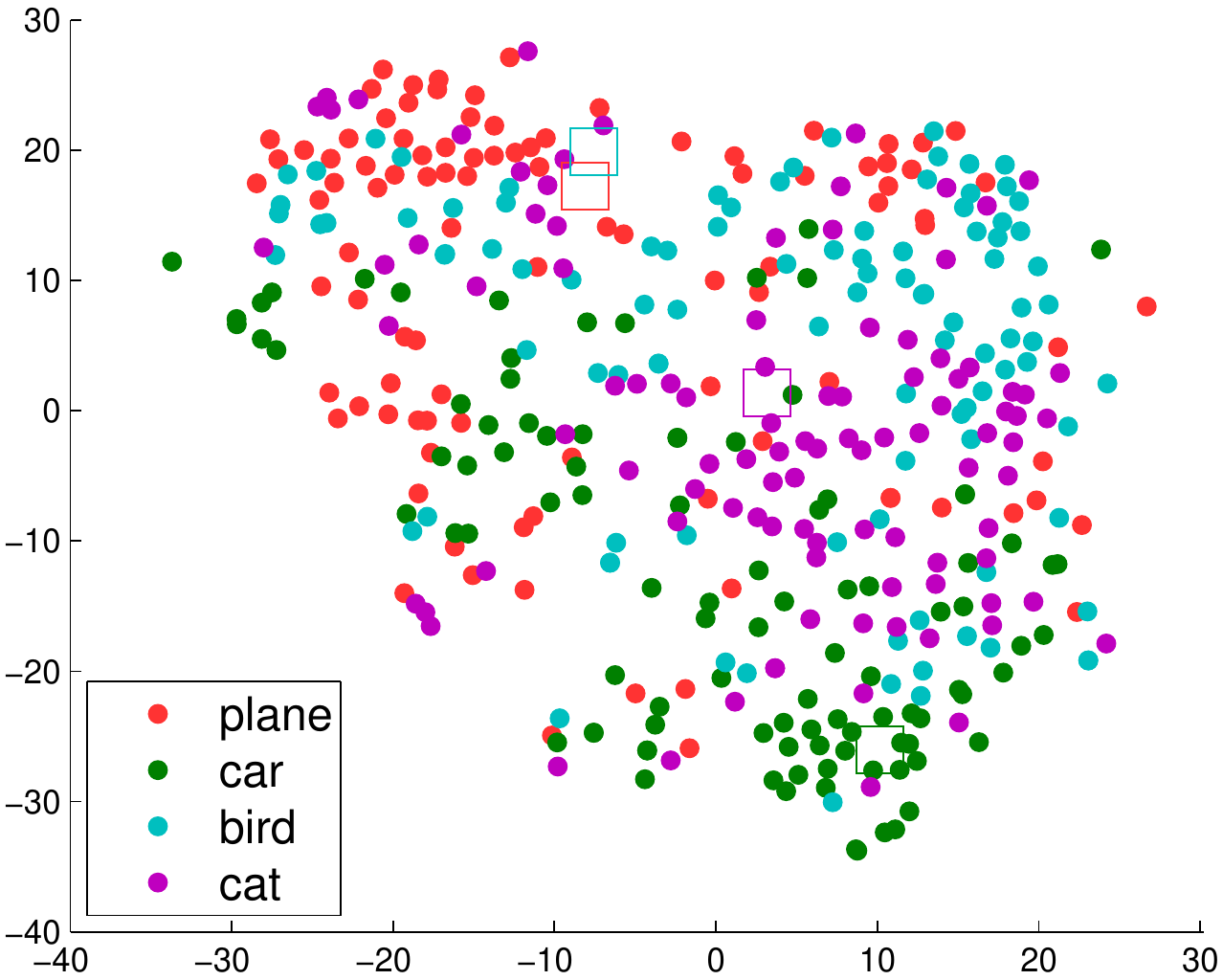}\label{fig:vizclustertask3}}\vspace{-0.4cm}
		\caption{Average classification accuracies over total 9 tasks (a) and 2D visualization of embedding clusters (solid circles) and prototypes (hollow squares) learnt by our method using t-sne \cite{van2014accelerating} on split CIFAR10 in incremental class task. Error bars in (a) denote standard errors over 10 runs. Refer to Sec.~\ref{subsec:baselines} for baseline method definitions. (b) The first task (Task = 1) is a binary classification problem. (c) The 3rd task is a 1-choose-4 classification problem. Colors in (b) and (c) correspond with the object classes in the legend. \vspace{-4mm}}
		\label{fig:quantativefewshots}
	\end{figure*}

\begin{figure*}[ht]
		\begin{center}
			\includegraphics[width=15cm]{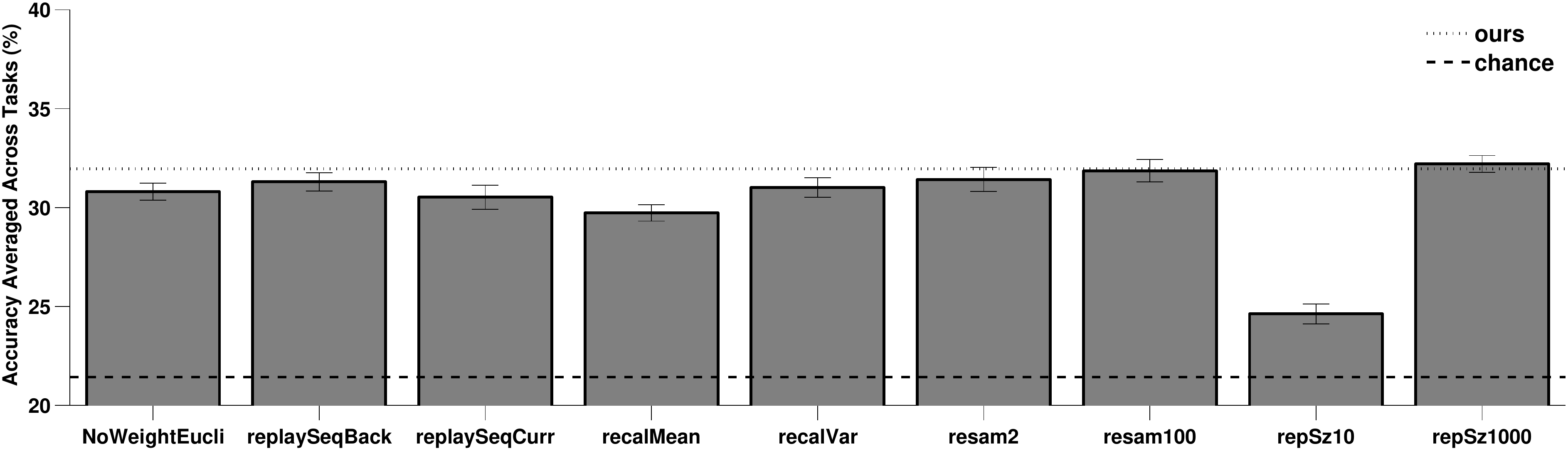}
		\end{center}
		\caption{Averaged classification accuracies over tasks for the ablated few-shot continual learning methods after 10 repeated runs on split CIFAR10 in incremental class task. See Sec.~\ref{subsec:ablation} for detailed description of each ablated method.\vspace{-4mm}}\vspace{-4mm}\vspace{-3mm}
		\label{fig:ablation}
\end{figure*}

\begin{figure*}[ht]
		\centering		
		\subfloat[Trajectories of prototypical means]{\includegraphics[width= 5.5cm, height = 5.0cm]{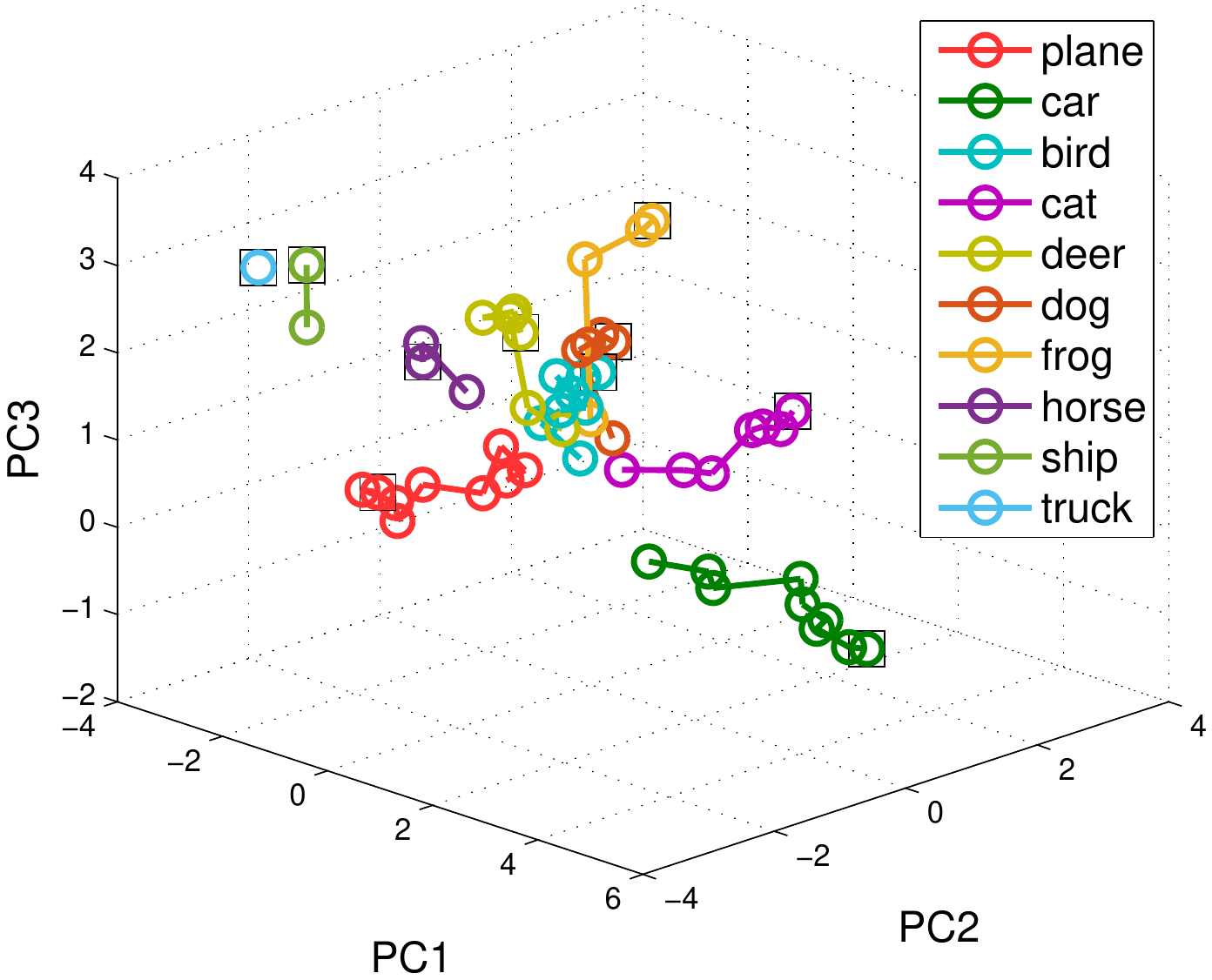}\label{fig:prototrajectory}}\hspace{0.4cm}
\subfloat[Feature similarity matrix]{\includegraphics[width= 5.3cm, height = 5.0cm]{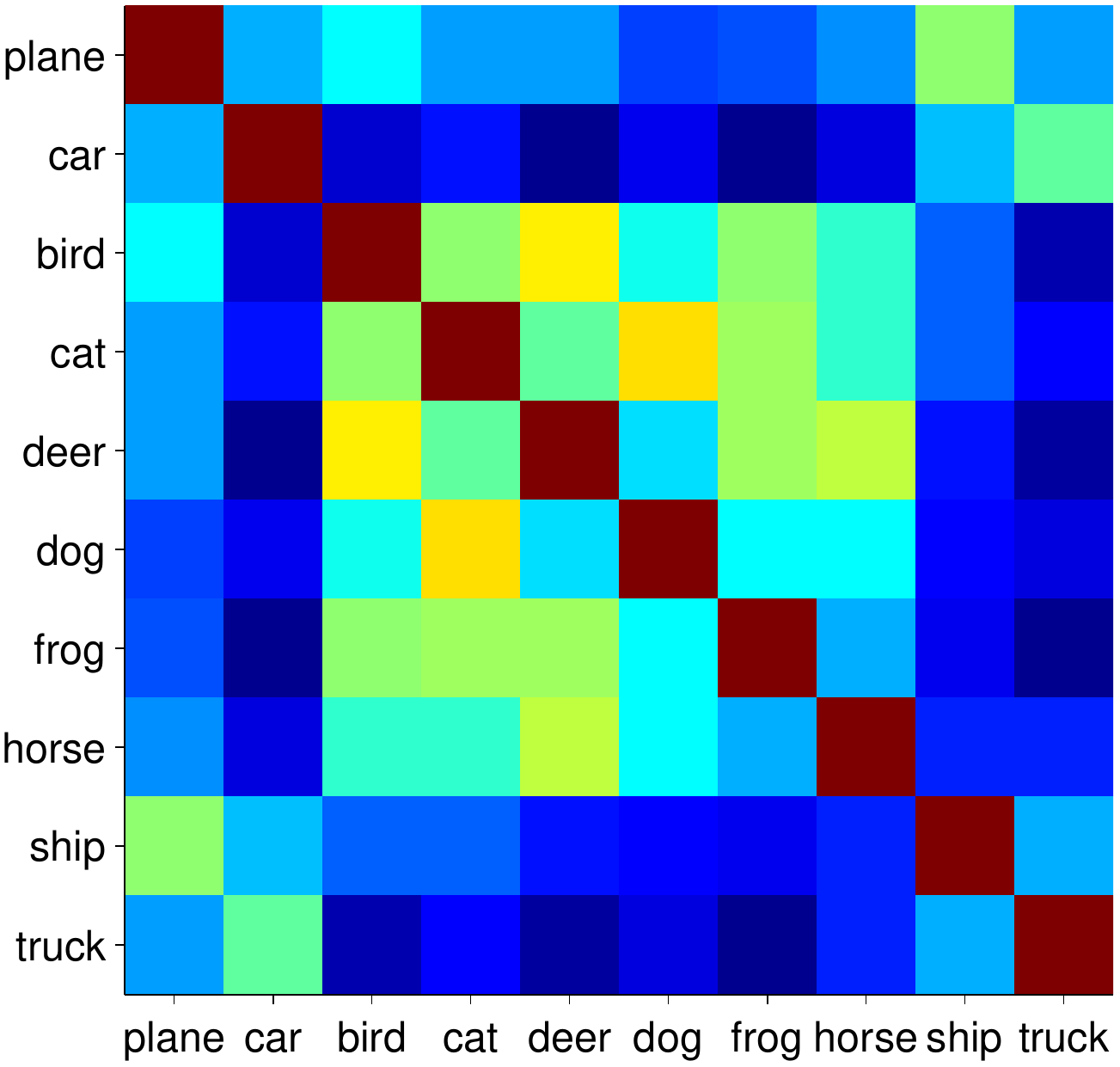}\label{fig:featureconfmat}}\hspace{0.3cm}
		\subfloat[Prototype dynamics similarity]{\includegraphics[width= 5.3cm, height = 5.0cm]{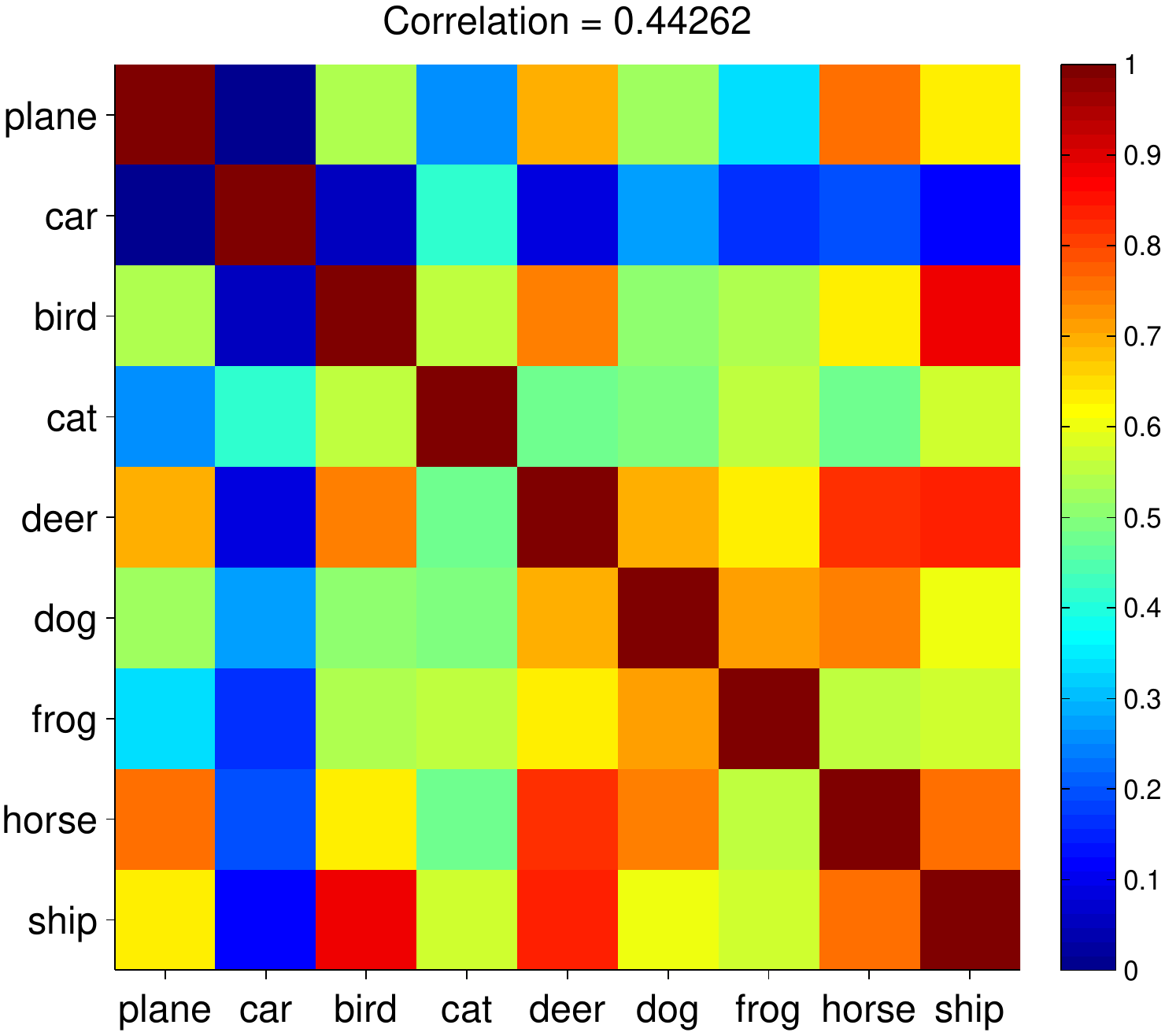}\label{fig:protodynamicsconfmat}}\hspace{0.2cm}
		\caption{Prototype dynamics analysis in split CIFAR10 in incremental class protocol. (a) Visualization of trajectory of prototypical mean movement over all 9 tasks after projecting all prototypical means of all tasks into 3D using first three principal components obtained from the latent feature space learnt in Task 1. The black squares denote the prototypical mean in the most recent tasks. The color code corresponds with object classes. (b) Feature similarity matrix is calculated using Euclidean distance between feature vectors extracted from second last layers of VGG16 network~\cite{simonyan2014very} pre-trained on ImageNet~\cite{deng2009imagenet}, which presumably ``saw" all the classes from CIFAR10 at once without any incremental class training. (c) For each class, a motion vector is calculated between the initial prototypical mean and the most recent prototypical mean across tasks. We can then compute the Euclidean distance for each pairs of motion vectors of the prototypical means from different classes. Refer to the colorbar on the right for similarity values. Correlation value reports the Pearson-correlation between feature similarity and prototype motion similarity. In other words, higher correlation values indicate that the more visually similar the two classes are; the prototypical means of these classes tend to move along in incremental class protocol.\vspace{-4mm}}\vspace{-4mm}\label{fig:protodynamicsanalysis}
	\end{figure*}
\subsection{Baselines}\label{subsec:baselines}
	
We include the following categories of continual learning methods for comparing with our method. To eliminate the effect of network structures in performance, we introduce control conditions with the same architecture complexity for all the methods in the same task across all the experiments except for the last layer before the softmax layer for classification. See Table~\ref{tab:memallocationCIFAR} for network architecture comparisons between baseline methods and our method.
	
\textbf{Parameter Regularization Methods}: Elastic Weight Consolidation (EWC) \cite{kirkpatrick2017overcoming}, Synaptic Intelligence (SI) \cite{zenke2017continual} and Memory Aware Synapses (MAS) \cite{aljundi2018memory}, where regularization terms are added in the loss function; online EWC \cite{kirkpatrick2017overcoming} which is an extension of EWC with scalability to a large number of tasks; L2 distance indicating parameter changes between tasks is added in the loss~\cite{kirkpatrick2017overcoming}; SGD, which is a naive baseline without any regularization terms, is optimized with Stochastic Gradient Descent \cite{bottou2010large}, sequentially over all tasks.
	
\textbf{Memory Distillation and Replay Methods}: incremental Classifier and Representation Learner (iCARL)~\cite{rebuffi2017icarl} proposes to regularize network behaviors by exact exemplar rehearsals via distillation loss.
	
Performance is reported in terms of both mean and standard deviation after 10 runs per protocol. Since generative model-based approaches \cite{van2018generative,shin2017continual} greatly alter architecture of the classification networks, we do not compare with them.

\subsection{Memory Comparison}\label{subsec:memorycomp}

For fair comparison, we compute the total number of parameters in a network for all the methods and allocate a comparable amount of memory as EWC \cite{kirkpatrick2017overcoming} and other parameter regularization methods, for storing example images per class and their variational prototypes in previous tasks. In EWC, the model allocates a memory size twice as the number of network parameters for computing the Fisher information matrix which is used for regularizing changes of network parameters \cite{kirkpatrick2017overcoming}. In more challenging classification tasks, the network size tends to be larger and hence, these methods require much more memory.

In Table~\ref{tab:memallocationCIFAR}, we show an example of memory allocation on split CIFAR10 in incremental class tasks. The feed-forward classification network used in baseline methods contains around $16.3\times 10^5$ parameters. Weight regularization methods require memory allocation twice as large, i.e., about $32.63\times 10^5$ parameters. The input RGB images are of size $3\times 32 \times 32$ and the variational prototypes contain one mean vector of size $1\times500$ and one variance vector of size $1\times500$. In example replay, we only store 1 example image and 1 variational prototype per class from previous tasks. The episodic memory of our method stores 10 images and 10 variational prototypes in total for all 10 classes, resulting in $21.7\times 10^5$ memory usage, which is 33\% less than weight regularization methods.
	
\vspace{-4mm}
\section{Results and Discussion}

In the main text, we focus on the results of our method in Split CIFAR10 in the incremental task. The Supp. Material shows results and discussion in the other two task protocols: permuted MNIST in incremental domain and split MiniImageNet in incremental class.

\subsection{Alleviating Forgetting}
	
Figure~\ref{fig:fewshotsavgcifar} reports the results of continual learning methods on split CIFAR10 in incremental class protocol. Our method (red) achieves the highest average classification accuracy among all the compared methods with minimum forgetting. Initially all compared continual learning methods outperform chance (dash line). Note that the chance is 1/2 in the first task. However, given 10 training samples in the subsequent tasks, all these algorithms except for L2 essentially fall to chance levels and fail to adapt to new tasks due to overfitting. A good continual learning method should not only show good memory retention but also be able to adapt to new tasks. Our method (red) consistently outperforms L2 across all tasks with an average improvement of 2.5\%. This reveals that our method performs classification via example replays and variational prototype regression in a more effective few-shot manner. Instead of replaying the exact prototypes of replayed images, the network is trained based on multiple samples from the distribution of $(\mu_{t,c}^n,\sigma_{t,c}^n)$ by finding their nearest variational prototypes per class. We also verified the importance of learning variational prototypes compared with only learning the prototypical mean vectors as shown in the ablation studies. Another advantage of our method over the others is that our network architecture is not dependent on the number of output classes and the knowledge in previous tasks can be well preserved and transferred. In traditional classification networks, new parameters often have to be added in the last classification layer as the total number of classes increases with increased numbers of tasks, which may easily lead to overfitting.

Fig.~\ref{fig:vizclustertask1} and Fig.~\ref{fig:vizclustertask3} provide visualizations of learnt variational means of image samples and variational prototypes from each class by projecting these latent representations into 2D space via the t-sne unsupervised dimension reduction method  \cite{van2014accelerating}. Given only 10 example images per class per task, our method is capable of clustering variational means of example images belonging to the same class and predicting their corresponding variational prototypes approximately in the center of each cluster. Over sequential tasks, our method accommodates new classes while maintaining the clustering of previous classes. From Task 1 to Task 3, our method incrementally learns two new classes (bird in blue and cat in purple) while latent representations of each image sample from two previous clusters (plane in red and car in green) remain clustered. However, in these two plots, the network parameters $F_t$ change across tasks; hence, the clusters of previous classes change in the latent feature space. In Sec.~\ref{subsec:prototypedynamics}, we analyze the dynamics of how the previous variational prototypes move in the original prototypical space learnt by $F_1$ with increasing number of new classes.

\subsection{Ablation Study}\label{subsec:ablation}

Here we assess the importance of several design choices in our method. Figure~\ref{fig:ablation} reports the classification accuracy of each ablated method averaged over all tasks.

First, in Equation~\ref{equ:prototype6}, the variance provides a ``confidence" measure of how good the learnt prototype mean is. Here, we replace the variance weighted Euclidean distance with a uniformly weighted Euclidean distance (NoWeightEucli) for nearest prototype classification loss. Compared with our proposed method, there is a moderate drop of 1.2\% in the average classification performance.

To prevent catastrophic forgetting, we replay stored example images and regresses the newly predicted variational distributions to be close to the stored prototype distribution for all previous tasks. We probe whether the sequence of replaying the variational prototypes from first task to recent ones matters. Replaying variational prototypes from the most recent tasks (replaySeqBack) results in 1\% performance drop and only replaying the most recent prototypes (replaySeqCurr) leads to further 0.3\% performance drop. Furthermore, we also analyze the effect of relaxing the mean and variance constraints. In other words, if the prototype recall only involves being close to a prototype mean (recalMean) or following a distribution with a similar prototype variance (recalVar), the performance is much worse than when combining both the mean and variance. This emphasizes the advantage of learning prototype distributions rather than a single prototype for a particular class in retaining memory of the previous tasks.

Next, to perform nearest prototype classification, we randomly sample multiple latent representations from the variational prototypes as shown in Fig~\ref{fig:model}. In our proposed method, we sample 50 variational prototypes. In the ablated methods, we titrate the sample size from 100 down to 2. Increasing sample sizes further from 50 to 100 saturates the performance; however, reducing sample sizes to 2 hinders the average classification accuracy by 0.5\%. We also vary the size of the variational prototype mean and variance. Increasing the latent feature space dimension from 500 to 1000 (repSz1000) boosts accuracy by 0.7\%; and vice versa for reducing the latent feature space (repSz10).

\subsection{Prototype Dynamics across Tasks}\label{subsec:prototypedynamics}

In split CIFAR10 in incremental class protocol, the network constantly updates its parameters from $F_1$ to $F_9$ over the total of 9 tasks. We report how the prototype means of previous classes change across tasks in Fig~\ref{fig:protodynamicsanalysis}. The visualization of the trajectory of prototype means across tasks in Fig.~\ref{fig:prototrajectory} suggests that, as the network incrementally learns more classes, the prototype means from previous classes move away from the center. To quantitatively measure how the visual feature similarity influence the prototype dynamics, we provide the visual feature similarity matrix in Fig.~\ref{fig:featureconfmat} and prototype dynamics similarity in Fig.~\ref{fig:protodynamicsconfmat}. A high correlation of 0.44 between feature similarity and prototype dynamics suggests that the dynamics of how the prototypes of two classes move is highly correlated with the visual feature similarities of these two classes. This observation provides some insights about how a classification network with our proposed method evolves a topological structure for learning to classify new objects in new tasks while keeping the previous classes separated apart from one another.
\vspace{-4mm}
\section{Conclusion}

We address the problem of catastrophic forgetting by proposing variational prototype replays in classification tasks. In addition to significantly alleviating catastrophic forgetting on benchmark datasets, our method is superior to others in terms of making the memory usage efficient, and being generalizable to learning novel concepts given only a few training examples in new tasks.

\bibliography{n_2019}

\begin{thebibliography}{37}
\providecommand{\natexlab}[1]{#1}
\providecommand{\url}[1]{\texttt{#1}}
\expandafter\ifx\csname urlstyle\endcsname\relax
  \providecommand{\doi}[1]{doi: #1}\else
  \providecommand{\doi}{doi: \begingroup \urlstyle{rm}\Url}\fi

\bibitem[Aljundi et~al.(2018)Aljundi, Babiloni, Elhoseiny, Rohrbach, and
  Tuytelaars]{aljundi2018memory}
Aljundi, R., Babiloni, F., Elhoseiny, M., Rohrbach, M., and Tuytelaars, T.
\newblock Memory aware synapses: Learning what (not) to forget.
\newblock In \emph{Proceedings of the European Conference on Computer Vision
  (ECCV)}, pp.\  139--154, 2018.

\bibitem[Atkinson et~al.(2018)Atkinson, McCane, Szymanski, and
  Robins]{atkinson2018pseudo}
Atkinson, C., McCane, B., Szymanski, L., and Robins, A.
\newblock Pseudo-recursal: Solving the catastrophic forgetting problem in deep
  neural networks.
\newblock \emph{arXiv preprint arXiv:1802.03875}, 2018.

\bibitem[Ba \& Caruana(2014)Ba and Caruana]{ba2014deep}
Ba, J. and Caruana, R.
\newblock Do deep nets really need to be deep?
\newblock In \emph{Advances in neural information processing systems}, pp.\
  2654--2662, 2014.

\bibitem[Bottou(2010)]{bottou2010large}
Bottou, L.
\newblock Large-scale machine learning with stochastic gradient descent.
\newblock In \emph{Proceedings of COMPSTAT'2010}, pp.\  177--186. Springer,
  2010.

\bibitem[Brahma \& Othon(2018)Brahma and Othon]{brahma2018subset}
Brahma, P.~P. and Othon, A.
\newblock Subset replay based continual learning for scalable improvement of
  autonomous systems.
\newblock In \emph{2018 IEEE/CVF Conference on Computer Vision and Pattern
  Recognition Workshops (CVPRW)}, pp.\  1179--11798. IEEE, 2018.

\bibitem[Bucilua et~al.(2006)Bucilua, Caruana, and
  Niculescu-Mizil]{bucilua2006model}
Bucilua, C., Caruana, R., and Niculescu-Mizil, A.
\newblock Model compression.
\newblock In \emph{Proceedings of the 12th ACM SIGKDD international conference
  on Knowledge discovery and data mining}, pp.\  535--541. ACM, 2006.

\bibitem[Chen et~al.(2012)Chen, Welling, and Smola]{chen2012super}
Chen, Y., Welling, M., and Smola, A.
\newblock Super-samples from kernel herding.
\newblock \emph{arXiv preprint arXiv:1203.3472}, 2012.

\bibitem[Deng et~al.(2009)Deng, Dong, Socher, Li, Li, and
  Fei-Fei]{deng2009imagenet}
Deng, J., Dong, W., Socher, R., Li, L.-J., Li, K., and Fei-Fei, L.
\newblock Imagenet: A large-scale hierarchical image database.
\newblock In \emph{2009 IEEE conference on computer vision and pattern
  recognition}, pp.\  248--255. Ieee, 2009.

\bibitem[Deng(2012)]{deng2012mnist}
Deng, L.
\newblock The mnist database of handwritten digit images for machine learning
  research [best of the web].
\newblock \emph{IEEE Signal Processing Magazine}, 29\penalty0 (6):\penalty0
  141--142, 2012.

\bibitem[Doersch(2016)]{doersch2016tutorial}
Doersch, C.
\newblock Tutorial on variational autoencoders.
\newblock \emph{arXiv preprint arXiv:1606.05908}, 2016.

\bibitem[French(1999)]{french1999catastrophic}
French, R.~M.
\newblock Catastrophic forgetting in connectionist networks.
\newblock \emph{Trends in cognitive sciences}, 3\penalty0 (4):\penalty0
  128--135, 1999.

\bibitem[Gidaris \& Komodakis(2018)Gidaris and Komodakis]{gidaris2018dynamic}
Gidaris, S. and Komodakis, N.
\newblock Dynamic few-shot visual learning without forgetting.
\newblock In \emph{Proceedings of the IEEE Conference on Computer Vision and
  Pattern Recognition}, pp.\  4367--4375, 2018.

\bibitem[Hassabis et~al.(2017)Hassabis, Kumaran, Summerfield, and
  Botvinick]{hassabis2017neuroscience}
Hassabis, D., Kumaran, D., Summerfield, C., and Botvinick, M.
\newblock Neuroscience-inspired artificial intelligence.
\newblock \emph{Neuron}, 95\penalty0 (2):\penalty0 245--258, 2017.

\bibitem[He \& Jaeger(2018)He and Jaeger]{he2018overcoming}
He, X. and Jaeger, H.
\newblock Overcoming catastrophic interference using conceptor-aided
  backpropagation.
\newblock 2018.

\bibitem[Hinton et~al.(2015)Hinton, Vinyals, and Dean]{hinton2015distilling}
Hinton, G., Vinyals, O., and Dean, J.
\newblock Distilling the knowledge in a neural network.
\newblock \emph{arXiv preprint arXiv:1503.02531}, 2015.

\bibitem[Kemker \& Kanan(2017)Kemker and Kanan]{kemker2017fearnet}
Kemker, R. and Kanan, C.
\newblock Fearnet: Brain-inspired model for incremental learning.
\newblock \emph{arXiv preprint arXiv:1711.10563}, 2017.

\bibitem[Kemker et~al.(2018)Kemker, McClure, Abitino, Hayes, and
  Kanan]{kemker2018measuring}
Kemker, R., McClure, M., Abitino, A., Hayes, T.~L., and Kanan, C.
\newblock Measuring catastrophic forgetting in neural networks.
\newblock In \emph{Thirty-second AAAI conference on artificial intelligence},
  2018.

\bibitem[Kirkpatrick et~al.(2017)Kirkpatrick, Pascanu, Rabinowitz, Veness,
  Desjardins, Rusu, Milan, Quan, Ramalho, Grabska-Barwinska,
  et~al.]{kirkpatrick2017overcoming}
Kirkpatrick, J., Pascanu, R., Rabinowitz, N., Veness, J., Desjardins, G., Rusu,
  A.~A., Milan, K., Quan, J., Ramalho, T., Grabska-Barwinska, A., et~al.
\newblock Overcoming catastrophic forgetting in neural networks.
\newblock \emph{Proceedings of the national academy of sciences}, 114\penalty0
  (13):\penalty0 3521--3526, 2017.

\bibitem[Koh \& Liang(2017)Koh and Liang]{koh2017understanding}
Koh, P.~W. and Liang, P.
\newblock Understanding black-box predictions via influence functions.
\newblock In \emph{Proceedings of the 34th International Conference on Machine
  Learning-Volume 70}, pp.\  1885--1894. JMLR. org, 2017.

\bibitem[Krizhevsky \& Hinton(2009)Krizhevsky and
  Hinton]{krizhevsky2009learning}
Krizhevsky, A. and Hinton, G.
\newblock Learning multiple layers of features from tiny images.
\newblock Technical report, Citeseer, 2009.

\bibitem[Lee et~al.(2017)Lee, Kim, Jun, Ha, and Zhang]{lee2017overcoming}
Lee, S.-W., Kim, J.-H., Jun, J., Ha, J.-W., and Zhang, B.-T.
\newblock Overcoming catastrophic forgetting by incremental moment matching.
\newblock In \emph{Advances in neural information processing systems}, pp.\
  4652--4662, 2017.

\bibitem[Li \& Hoiem(2018)Li and Hoiem]{li2018learning}
Li, Z. and Hoiem, D.
\newblock Learning without forgetting.
\newblock \emph{IEEE transactions on pattern analysis and machine
  intelligence}, 40\penalty0 (12):\penalty0 2935--2947, 2018.

\bibitem[Lopez-Paz et~al.(2017)]{lopez2017gradient}
Lopez-Paz, D. et~al.
\newblock Gradient episodic memory for continual learning.
\newblock In \emph{Advances in Neural Information Processing Systems}, pp.\
  6467--6476, 2017.

\bibitem[Maltoni \& Lomonaco(2019)Maltoni and Lomonaco]{maltoni2019continuous}
Maltoni, D. and Lomonaco, V.
\newblock Continuous learning in single-incremental-task scenarios.
\newblock \emph{Neural Networks}, 2019.

\bibitem[McCloskey \& Cohen(1989)McCloskey and
  Cohen]{mccloskey1989catastrophic}
McCloskey, M. and Cohen, N.~J.
\newblock Catastrophic interference in connectionist networks: The sequential
  learning problem.
\newblock In \emph{Psychology of learning and motivation}, volume~24, pp.\
  109--165. Elsevier, 1989.

\bibitem[Nguyen et~al.(2017)Nguyen, Li, Bui, and Turner]{nguyen2017variational}
Nguyen, C.~V., Li, Y., Bui, T.~D., and Turner, R.~E.
\newblock Variational continual learning.
\newblock \emph{arXiv preprint arXiv:1710.10628}, 2017.

\bibitem[Ratcliff(1990)]{ratcliff1990connectionist}
Ratcliff, R.
\newblock Connectionist models of recognition memory: constraints imposed by
  learning and forgetting functions.
\newblock \emph{Psychological review}, 97\penalty0 (2):\penalty0 285, 1990.

\bibitem[Rebuffi et~al.(2017)Rebuffi, Kolesnikov, Sperl, and
  Lampert]{rebuffi2017icarl}
Rebuffi, S.-A., Kolesnikov, A., Sperl, G., and Lampert, C.~H.
\newblock icarl: Incremental classifier and representation learning.
\newblock In \emph{Proceedings of the IEEE Conference on Computer Vision and
  Pattern Recognition}, pp.\  2001--2010, 2017.

\bibitem[Robins(1995)]{robins1995catastrophic}
Robins, A.
\newblock Catastrophic forgetting, rehearsal and pseudorehearsal.
\newblock \emph{Connection Science}, 7\penalty0 (2):\penalty0 123--146, 1995.

\bibitem[Shin et~al.(2017)Shin, Lee, Kim, and Kim]{shin2017continual}
Shin, H., Lee, J.~K., Kim, J., and Kim, J.
\newblock Continual learning with deep generative replay.
\newblock In \emph{Advances in Neural Information Processing Systems}, pp.\
  2990--2999, 2017.

\bibitem[Simonyan \& Zisserman(2014)Simonyan and Zisserman]{simonyan2014very}
Simonyan, K. and Zisserman, A.
\newblock Very deep convolutional networks for large-scale image recognition.
\newblock \emph{arXiv preprint arXiv:1409.1556}, 2014.

\bibitem[Snell et~al.(2017)Snell, Swersky, and Zemel]{snell2017prototypical}
Snell, J., Swersky, K., and Zemel, R.
\newblock Prototypical networks for few-shot learning.
\newblock In \emph{Advances in Neural Information Processing Systems}, pp.\
  4077--4087, 2017.

\bibitem[Thrun \& Mitchell(1995)Thrun and Mitchell]{thrun1995lifelong}
Thrun, S. and Mitchell, T.~M.
\newblock Lifelong robot learning.
\newblock \emph{Robotics and autonomous systems}, 15\penalty0 (1-2):\penalty0
  25--46, 1995.

\bibitem[van~de Ven \& Tolias(2018)van~de Ven and Tolias]{van2018generative}
van~de Ven, G.~M. and Tolias, A.~S.
\newblock Generative replay with feedback connections as a general strategy for
  continual learning.
\newblock \emph{arXiv preprint arXiv:1809.10635}, 2018.

\bibitem[Van Der~Maaten(2014)]{van2014accelerating}
Van Der~Maaten, L.
\newblock Accelerating t-sne using tree-based algorithms.
\newblock \emph{The Journal of Machine Learning Research}, 15\penalty0
  (1):\penalty0 3221--3245, 2014.

\bibitem[Wen et~al.(2018)Wen, Cao, and Huang]{wen2018few}
Wen, J., Cao, Y., and Huang, R.
\newblock Few-shot self reminder to overcome catastrophic forgetting.
\newblock \emph{arXiv preprint arXiv:1812.00543}, 2018.

\bibitem[Zenke et~al.(2017)Zenke, Poole, and Ganguli]{zenke2017continual}
Zenke, F., Poole, B., and Ganguli, S.
\newblock Continual learning through synaptic intelligence.
\newblock In \emph{Proceedings of the 34th International Conference on Machine
  Learning-Volume 70}, pp.\  3987--3995. JMLR. org, 2017.

\end{thebibliography}
\bibliographystyle{icml2020}

\end{document}